\documentclass{article}

\usepackage{microtype}
\usepackage{graphicx}
\usepackage{subcaption}
\usepackage{booktabs}
\usepackage{multirow}
\usepackage{hyperref}
\usepackage{amsmath}
\usepackage{amssymb}
\usepackage{mathtools}
\usepackage[capitalize,noabbrev]{cleveref}
\usepackage{dblfloatfix}
\usepackage{float}

\usepackage[accepted]{icml2026}

\makeatletter
\@namedef{ver@inputenc.sty}{prevented}
\makeatother

\usepackage{fontspec}

\usepackage[english]{babel}
\babelprovide[import, onchar=ids fonts]{arabic}
\babelprovide[import, onchar=ids fonts]{hindi}
\babelprovide[import, onchar=ids fonts]{russian}

\babelfont{rm}[
  BoldFont = texgyretermes-bold.otf,
  ItalicFont = texgyretermes-italic.otf,
  BoldItalicFont = texgyretermes-bolditalic.otf
]{texgyretermes-regular.otf}

\babelfont[arabic]{rm}[
  BoldFont = Amiri-Bold.ttf,
  ItalicFont = Amiri-Italic.ttf,
  BoldItalicFont = Amiri-BoldItalic.ttf
]{Amiri-Regular.ttf}

\babelfont[hindi]{rm}[
  BoldFont = Eczar-Bold.otf
]{Eczar-Regular.otf}

\babelfont[russian]{rm}[
  BoldFont = DejaVuSerif-Bold.ttf,
  ItalicFont = DejaVuSerif-Italic.ttf,
  BoldItalicFont = DejaVuSerif-BoldItalic.ttf
]{DejaVuSerif.ttf}

\usepackage{xeCJK}
\icmltitlerunning{Personalization, Personas, and Forecasting in Value Alignment}

\begin{document}

\twocolumn[
\icmltitle{Personalization, Personas, and Forecasting in Value Alignment}

\begin{icmlauthorlist}
\icmlauthor{James Wedgwood}{cmu}
\icmlauthor{Pratiksha Thaker}{cmu}
\icmlauthor{Neil Kale}{cmu}
\icmlauthor{Virginia Smith}{cmu}
\end{icmlauthorlist}

\icmlaffiliation{cmu}{Carnegie Mellon University, Pittsburgh, PA, USA}

\icmlcorrespondingauthor{James Wedgwood}{jwedgwoo@andrew.cmu.edu}
\icmlcorrespondingauthor{Pratiksha Thaker}{pthaker@andrew.cmu.edu}
\icmlcorrespondingauthor{Neil Kale}{nkale@andrew.cmu.edu}
\icmlcorrespondingauthor{Virginia Smith}{smithv@andrew.cmu.edu}

\icmlkeywords{large language models, cultural modeling, survey response prediction, World Values Survey}

\vskip 0.3in
]

\printAffiliationsAndNotice{}

\begin{abstract}
LLM behavior may be conditioned by human identity in several ways: they may be asked to adapt to users, role-play populations, or forecast how people would answer value-laden questions. We test whether these framings are interchangeable using the World Values Survey (WVS). We evaluate GPT-5.4, Claude Sonnet 4.6, Gemini 2.5 Flash, and Qwen3-235B on 101 WVS-derived questions across 13 language-country slices, comparing a language-only baseline with user-country, persona-country, and third-person prompts. Across 21,008 model-response rows, prompt framing is a first-order determinant of cultural alignment: country cues often shift answers substantially, but not all shifts move toward matched human response distributions. Third-person forecasting yields the strongest directional alignment for three of the four hosted models, while personalization and role-play are weaker or less stable. Alignment gains concentrate on salient value dimensions such as religiosity, gender roles, and work-oriented material values, whereas institutional trust and democracy-related questions remain difficult. These results show that prompt framing is not a cosmetic choice in cultural value elicitation; it changes both model behavior and measured alignment.
\end{abstract}

\section{Introduction}
\label{sec:intro}

\begin{figure*}[!t]
\centering
\includegraphics[width=0.9\textwidth]{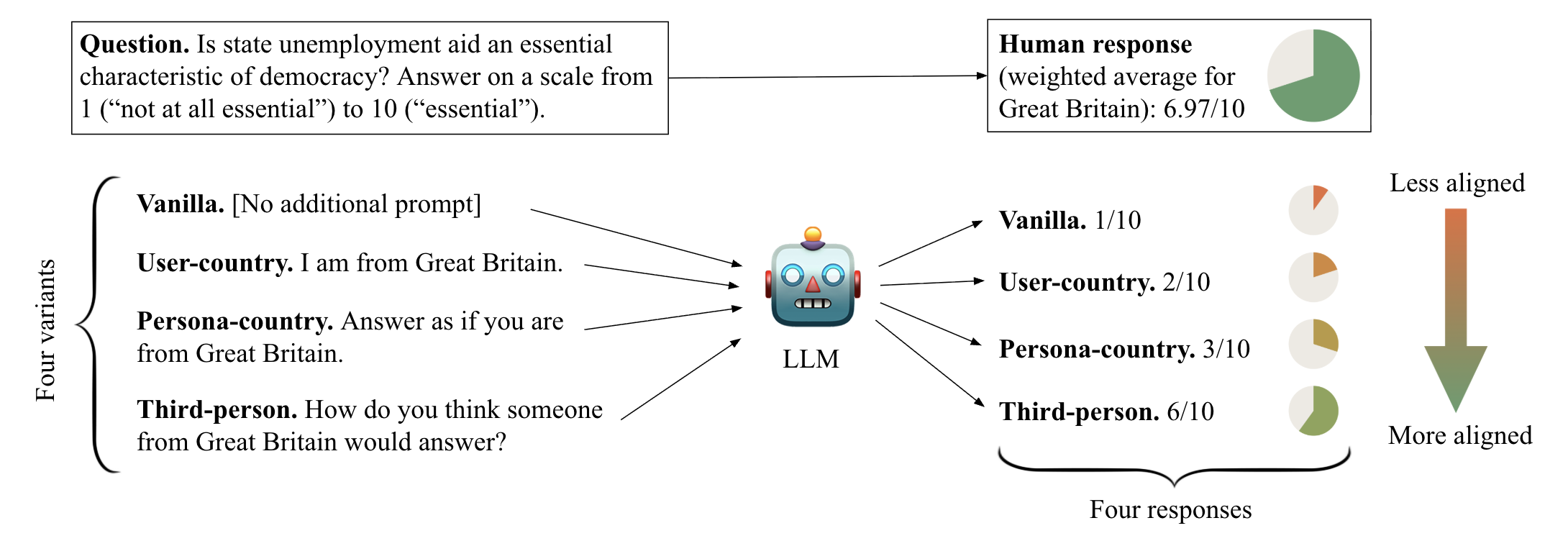}
\caption{An example of the four prompt conditions tested in this paper. This is a real example showing responses supplied by GPT-5.4 and highlighting the most common trend in condition-dependent directional alignment. The WVS question text was simplified for clarity.}
\label{fig:example}
\end{figure*}

Large language models are increasingly deployed across countries, languages, and cultural contexts, making it important to understand not only whether their responses reflect a broad range of human values, but also how those responses change when models are given information about human identity. This question matters both for practical deployment and for scientific evaluation: a model may be asked to adapt to an individual user, to simulate a population, or to predict how a person from a particular demographic group would answer. These settings are closely related, but they are not identical, and treating them as interchangeable risks obscuring important differences in model behavior.

In practical usage, one can broadly define three approaches to
shaping LLM behavior with awareness of a description of a human identity.
\textbf{Personalization} provides information about the user 
interacting with the LLM to improve the user experience,
either through a system prompt or through interactions with the user.
A \textbf{persona} sets up a role play exercise, encouraging the LLM to respond as if \emph{it is}
a human with a given set of traits;
this approach has been used, for example, in simulating population surveys, A/B tests, and focus groups. Finally, \textbf{forecasting} asks the LLM to predict how someone from the target demographic would respond without asking it explicitly to act as if it belongs to that group; this approach may be used to study how a particular identity category is perceived and what assumptions are likely to be made about its members.

Na\"ively, one might expect consistency between these three modalities.
Consider the following example:
an LLM with a persona of a woman aged 20 in the United States is queried ``are you old enough to drive?'';
a second LLM personalized to respond to a woman aged 20 in the United States is queried ``am I old enough to drive?'';
while a third LLM is asked simply ``is a woman aged 20 in the United States old enough to drive?''
Given the factual nature of this question,
all three should respond in the same way.

The expected outcomes are less clear in the context of
value-based questions that do not have factual answers,
when simulating populations or predicting the values of users.
While a significant amount of attention \citep{durmus2023_global_opinions, Tao_2024, alkhamissi2024investigatingculturalalignmentlarge}
has been devoted to independently studying LLM personalization, personas, and forecasting, 
to our knowledge there is no work that evaluates consistency
between the three.

In this work, we use the World Values Survey (WVS) \citep{haerpfer2020wvs}—the largest ever non-commercial, cross-national study of human beliefs and values—
as a basis for studying consistency between persona-based queries
and personalized queries to elicit value judgments from LLMs.
Our focus is on LLM modeling of the value systems of different countries, and
we study four settings: (1) \textbf{Vanilla/Language}: the survey question with no
additional descriptor information, translated when applicable into the language matched to a given country; 
(2) \textbf{User-country}: a prompt stating that the \emph{user} is from the given country;
(3) \textbf{Persona-country}: a prompt asking the \emph{model} to answer as though it were
a person from the given country;
and (4) \textbf{Third-person}: a prompt asking the model to predict how a person 
from the given country \emph{would answer}. An illustrative example of these four prompt conditions is shown in Figure~\ref{fig:example}.

It is not immediately clear what the desired behavior should be under these conditions. For example, user-country prompting does not explicitly ask the LLM to tailor its response, so one might expect that the responses should be the same as the vanilla case; on the other hand, a degree of conformity to perceived user opinions is likely desirable in some cases. Even if we believe that the answers should be adjusted here, it is unclear whether language prompting should also result in modified responses relative to the English vanilla baseline: does a user merely speaking a particular language imply something about their identity and values? Meanwhile, the persona-country and third-person prompts, despite their differences in framing, are both asking that the LLM guess the response of an individual from a given country. Should these responses be the same? 

Regardless of one's stance on these normative questions, several clear trends emerge in practice, largely consistently across the four models we survey. First, country-conditioned prompts generally move model responses more than language-only prompts, but this movement is not always aligned with the corresponding WVS response distribution. Second, explicit third-person forecasting produces the clearest movement toward the matched human target for three of the four hosted models, suggesting that models are better at predicting country-level survey responses when the task is framed as prediction rather than role play or personalization. Third, the gains are concentrated on socially legible value dimensions such as religiosity, gender roles, security tradeoffs, and work-oriented material values, while institutional trust and democracy-related questions remain harder.

Taken together, these results suggest marked differences among the three modalities of personalization, personas, and forecasting, which hold with remarkable consistency across model families. In view of these findings, we encourage future research into LLM biases and cultural modeling to proceed with explicit acknowledgment of these discrepancies and to select the modality most appropriate to the given setting in a careful and principled way.

\section{Related Work}

\paragraph{Stereotyping, flattening, and the limits of cultural simulation.}
Demographic-conditioned prompting can move model responses toward aggregate survey targets, but prior work cautions against interpreting such movement as faithful. \citet{durmus2023_global_opinions} find that cross-national prompting can elicit stereotyped justifications, while \citet{wang2025large} argue that replacing human participants with LLMs can misportray identity groups. We therefore treat country-conditioned prompting as a descriptive object of study rather than as a general solution for cultural representation.

\vspace{-0.15in}
\paragraph{Survey-based evaluations of LLM values and cultural alignment.}
A growing literature evaluates language models against structured social surveys, using instruments such as the World Values Survey (WVS), Pew surveys, and Hofstede-style cultural questionnaires. \citet{durmus2023_global_opinions} introduce GlobalOpinionQA using WVS and Pew questions, \citet{cao2023assessingcrossculturalalignmentchatgpt} assess ChatGPT against Hofstede's cultural dimensions, and \citet{Tao_2024} place GPT responses on the Inglehart--Welzel cultural map. \citet{alkhamissi2024investigatingculturalalignmentlarge} simulate WVS respondents from Egypt and the United States, while \citet{zhao2024worldvaluesbenchlargescalebenchmarkdataset} introduce WorldValuesBench for predicting WVS responses conditioned on demographic attributes. These studies typically choose a single elicitation frame rather than comparing the consequences of different frames.

\vspace{-0.15in}
\paragraph{Personalization, personas, and forecasting.}
Prior work has used all three identity-conditioning modalities considered here, often without making the choice of modality a central object of study. Some work frames identity as user context, corresponding to personalization or first-person prompting \citep{venkata2026whenai, khanuja2026steering}. Other work asks the model to adopt a persona: \citet{Tao_2024} prompt models as an ``average human being'' in a target country, \citet{alkhamissi2024investigatingculturalalignmentlarge} use demographic role-play prompts, and \citet{santurkar2023whose} study demographic opinion alignment under a PORTRAY condition. A third set of work instead frames the task as forecasting another respondent's answer: \citet{durmus2023_global_opinions} ask how someone from a target country would respond, \citet{zhao2024worldvaluesbenchlargescalebenchmarkdataset} ask what a described ``Person X'' would answer, and \citet{lutz2025promptmakespersona} compare several persona prompts. Our work treats these framing choices as experimentally distinct rather than interchangeable.

\section{Methods}
\subsection{Task Setting}
We study whether country-conditioned prompting moves LLM responses toward the response distributions observed for the corresponding country in the World Values Survey (WVS) \cite{haerpfer2020wvs}. WVS is a large-scale, cross-national survey program measuring people's social, political, moral, and cultural values across many countries over multiple waves; in this paper, we use the most recent wave as of the time of writing (2017–2022). WVS has been widely used in the AI literature \citep{durmus2023_global_opinions, zhao2024worldvaluesbenchlargescalebenchmarkdataset, Tao_2024, liu2025alignmentlargelanguagemodels} due to its comprehensive scope and because it provides a structured way to compare model behavior against human value distributions across cultures, rather than treating ``human preferences" as culturally uniform.

We evaluate four frontier models—GPT-5.4, Claude Sonnet 4.6, Gemini 2.5 Flash, and Qwen3-235B—on 101 WVS-derived questions across 13 language-country slices, selected to cover the world's most widely-spoken languages across countries with large populations of WVS respondents. The slices include seven interview languages and thirteen country conditions: Arabic (Egypt, Jordan, Morocco), English (Great Britain, Kenya, United States), Spanish (Argentina, Colombia, Mexico), French (Canada), Hindi (India), Russian (Russia), and Chinese (China). This yields 21,008 model-response rows across the sixteen model-prompt combinations.

\subsection{Question Set and Selection Units}
\label{sec:question-set}
The benchmark is built from a manually curated WVS subset of 101 \emph{standalone question units}, selected for cross-country portability and standalone interpretability. A large majority of these question units correspond to individual WVS question IDs, but there are several cases in which WVS indexes each one of a series of checklist-style answers as a separate question; in such cases, we grouped all answers together into a single unit. For ease of notation, we will refer to these derived units simply as ``questions."

In selecting questions, we excluded those whose wording includes country-specific party systems, local institution lists, subjective self-evaluation (for example, life satisfaction or self-rated health), or references to local neighborhoods and communities, as such items are not applicable to LLM respondents (i.e., in the persona setting). The admissible answers for each question may be binary (e.g. yes/no), ordered multiple choice (e.g. strongly agree / agree / disagree / strongly disagree), or unordered multiple choice. For more details on question definitions and selection, see Appendix~\ref{app:selection}.

To facilitate nuanced analysis, we divide the selected questions into seven mutually exclusive and collectively exhaustive categories: Morality and Worldview (24 questions); Democracy, Governance, and Public Authority (20); Trust, Civic Belonging, and Institutions (14); Economic Fairness, Work, and Material Conditions (13); Gender, Equality, and Social Inclusion (11); Religion and Science (10); and Public Order, Corruption, Migration, and Security (9).

\subsection{Prompt Conditions}
We evaluate four prompt conditions, as illustrated in Figure~\ref{fig:example}:
\begin{enumerate}
\item \textbf{Language}: a language-matched vanilla prompt with no country cue.
\vspace{-0.05in}
\item \textbf{User-country}: the prompt states that the user is from a specified country.
\vspace{-0.05in}
\item \textbf{Persona-country}: the model is asked to answer as if it were a person from that country.
\vspace{-0.05in}
\item \textbf{Third-person}: the model is asked to predict how a person from that country would answer.

\end{enumerate}

User-country, persona-country, and third-person prompting respectively proxy the personalization, persona, and forecasting modalities discussed in Section~\ref{sec:intro}. The user-country and persona-country variants still invite first-person answering, and the user-country prompt does not explicitly invite the model to personalize its answer. The third-person prompt, meanwhile, reframes the task as forecasting a respondent from the target country. Appendix~\ref{app:prompts} gives the verbatim prompt wording across the four variants for all seven languages.

\subsection{Human Targets and Response Distributions}
Each model response is compared to a matched human reference distribution derived from WVS responses. For vanilla prompts in language $\ell$, the human target is the weighted response distribution among respondents interviewed in $\ell$. For country-conditioned prompts in language-country slice $(\ell, c)$, the human target is the weighted response distribution among respondents interviewed in $\ell$ in country $c$.

For each question, the human response distribution is the normalized share of respondent weight assigned to each answer option, excluding codes indicating that the respondent did not give a substantive answer. For example, a value of $0.62$ on one answer option means that 62\% of substantive responses in the matched group selected that option.

\begin{figure*}[!t]

\centering
\small
\begin{tabular}{llrrrrr}
\toprule
Model & Prompt & Coverage & Baseline gap & Response gap & Shift & Align. \\
\midrule
GPT-5.4 & Language & 99.7\% & 0.337 & 0.341 & 0.137 & -0.005 \\
GPT-5.4 & User-country & 96.7\% & 0.343 & 0.336 & 0.166 & 0.007 \\
GPT-5.4 & Persona-country & 96.9\% & 0.344 & 0.313 & 0.193 & 0.031 \\
GPT-5.4 & Third-person & 96.7\% & 0.343 & 0.270 & 0.261 & 0.073 \\
\midrule
Claude Sonnet 4.6 & Language & 92.5\% & 0.309 & 0.319 & 0.105 & -0.010 \\
Claude Sonnet 4.6 & User-country & 91.6\% & 0.310 & 0.292 & 0.150 & 0.018 \\
Claude Sonnet 4.6 & Persona-country & 92.7\% & 0.311 & 0.291 & 0.192 & 0.019 \\
Claude Sonnet 4.6 & Third-person & 92.7\% & 0.312 & 0.293 & 0.209 & 0.019 \\
\midrule
Gemini 2.5 Flash & Language & 85.1\% & 0.359 & 0.348 & 0.128 & 0.012 \\
Gemini 2.5 Flash & User-country & 80.1\% & 0.362 & 0.324 & 0.168 & 0.038 \\
Gemini 2.5 Flash & Persona-country & 86.1\% & 0.368 & 0.309 & 0.216 & 0.059 \\
Gemini 2.5 Flash & Third-person & 79.0\% & 0.366 & 0.291 & 0.238 & 0.075 \\
\midrule
Qwen3-235B & Language & 99.1\% & 0.311 & 0.345 & 0.128 & -0.034 \\
Qwen3-235B & User-country & 96.6\% & 0.317 & 0.324 & 0.169 & -0.007 \\
Qwen3-235B & Persona-country & 96.6\% & 0.317 & 0.297 & 0.216 & 0.020 \\
Qwen3-235B & Third-person & 96.3\% & 0.317 & 0.283 & 0.255 & 0.034 \\
\bottomrule
\end{tabular}

\centering
\includegraphics[width=\textwidth]{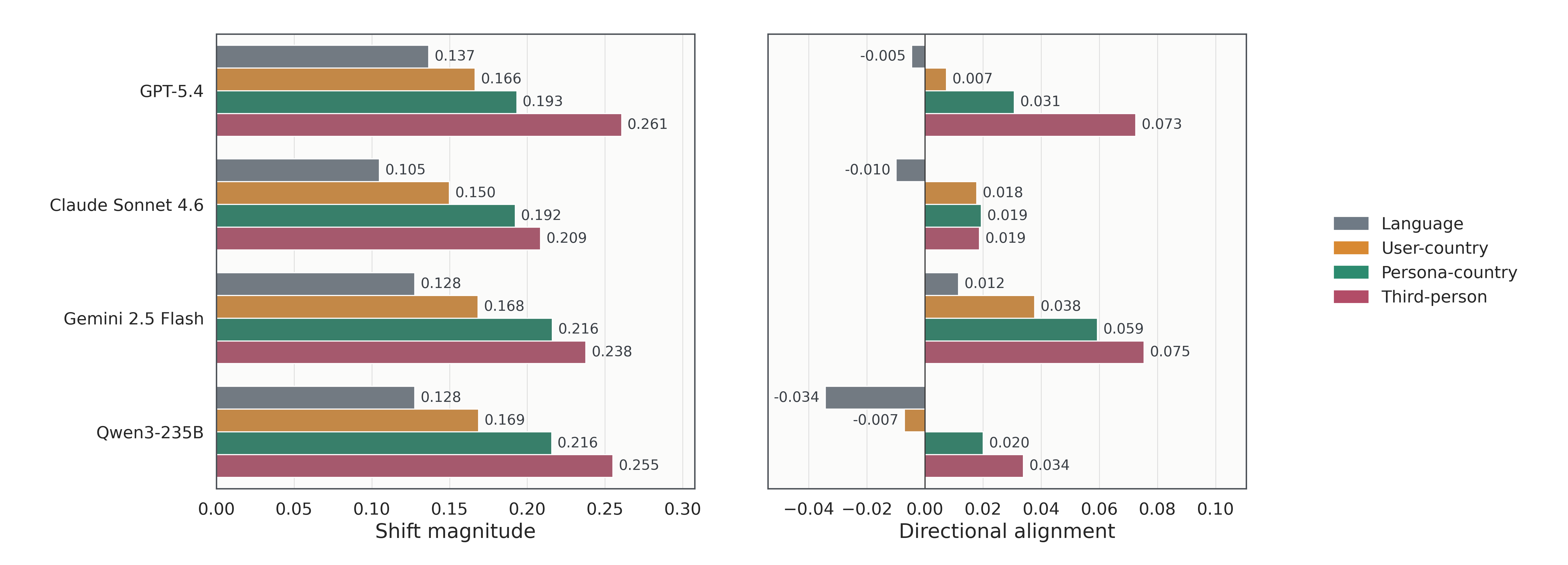}
\caption{Overall prompt-effect metrics for GPT-5.4, Claude Sonnet 4.6, Gemini 2.5 Flash, and Qwen3-235B. The chart provides a visual depiction of the final two columns of the table (shift magnitude and directional alignment), highlighting the Language < User-country < Persona-country < Third-person trend observed for all models on the shift magnitude metric, and for all models except Claude Sonnet 4.6 on the directional alignment metric.}
\label{fig:overall-metrics}
\end{figure*}

\subsection{Scoring}
Let $B(m, q)$ denote model $m$'s English-vanilla response for question $q$, which will be used throughout as a baseline representing the unconditioned ``opinion" of $m$ on $q$. Let $R(m, q, \ell, c, v)$ denote the response under language $\ell$, country $c$, and prompt variant $v$ and let $H(q, \ell, c)$ denote the matched human response distribution for language $\ell$ and country $c$. For language prompting, we do not condition on country and instead compare $R(m, q, \ell, v)$ to $H(q, \ell)$, the matched human response distribution for language $\ell$. We score responses with normalized distances on $[0,1]$:
\begin{align}
\text{baseline\_gap} &= d(B, H), \\
\text{response\_gap} &= d(R, H), \\
\text{shift\_magnitude} &= d(R, B), \\
\text{directional\_alignment} &= \text{baseline\_gap} - \text{response\_gap}.
\end{align}

Positive directional alignment means that prompting moved the response closer to the matched WVS distribution; negative values mean movement away from it. Shift magnitude is reported separately because a large change is not necessarily movement toward the human target. We refer to large positive directional alignment scores as ``strong" throughout this paper. However, it is worth noting that this terminology is descriptive and does not indicate a normative stance that high alignment is always positive under any prompt condition; such prescriptive questions are subjective and, as such, our position on them is agnostic.

For ordered questions we use normalized Wasserstein distance over ordered answer positions. For unordered single-choice questions we use total variation distance. We score bundled units on their member questions first and then average them back to the parent selection unit.

\subsection{Semantic Axes}
Beyond question-level distances, we analyze direction of movement on a fixed set of ten semantic axes: gender egalitarianism, outgroup inclusion, religiosity / faith primacy, science / technology optimism, traditional authority, civic integrity, economic interventionism, productivist materialism, institutional trust, and democratic pluralism. Note that although there is some conceptual overlap between these semantic axes and the seven categories from Section~\ref{sec:question-set}, they have different functions: the categories are used to segment questions for nuanced analysis, while the directionality of the semantic axes give a view into the tendencies of LLM versus human responses on broad issues, allowing us to make statements like ``GPT-5.4 systematically overestimates optimism about science and technology under Latin American country prompting." Unlike the categories, the semantic axes are not one-to-one with questions: of the 101 total questions used, 74 are mapped to exactly one axis, 14 are mapped to multiple axes, and 13 are not mapped.

For each mapped question $q$ on axis $a$, we define an oriented semantic score $s_q(\cdot) \in [0,1]$ such that larger values always indicate more of the positive pole of axis $a$. Ordered questions are rescaled to $[0,1]$ and reverse-coded when needed; binary member items are coded as 0/1 and oriented in the same way; and a small number of unordered items with clear practical ordering use an explicit hand coding from response option to semantic value. Details on these axes, the questions mapped to each of them, and the poles defined as ``positive" and ``negative" can be found in Appendix~\ref{app:axes}.

We then define
\begin{align}
\text{baseline\_semantic}(q,a) &= s_q(B), \\
\text{response\_semantic}(q,a) &= s_q(R), \\
\text{human\_semantic}(q,a) &= \mathbb{E}_{H}[s_q].
\end{align}
From these quantities we compute
\begin{align}
\text{semantic\_shift}(q,a) &= \nonumber \\ \text{response\_semantic}(q,a) - \text{baseline\_semantic}(q,a), \\
\text{semantic\_alignment\_delta}(q,a) &= \nonumber \\\left|\text{human\_semantic}(q,a) - \text{baseline\_semantic}(q,a)\right| - \nonumber \\ \left|\text{human\_semantic}(q,a) - \text{response\_semantic}(q,a)\right|.
\end{align}
Positive semantic shift means movement toward the positive pole of the axis. Positive semantic alignment delta means movement toward the matched human target on that axis, analogous to directional alignment in the question-scoring case. Axis-level summaries use the same weighting principle as the main benchmark: bundled members are scored first, averaged back to the parent selection unit, and then parent units are averaged with equal weight within each axis.

\begin{figure*}[!t]
\centering
\includegraphics[width=0.9\textwidth]{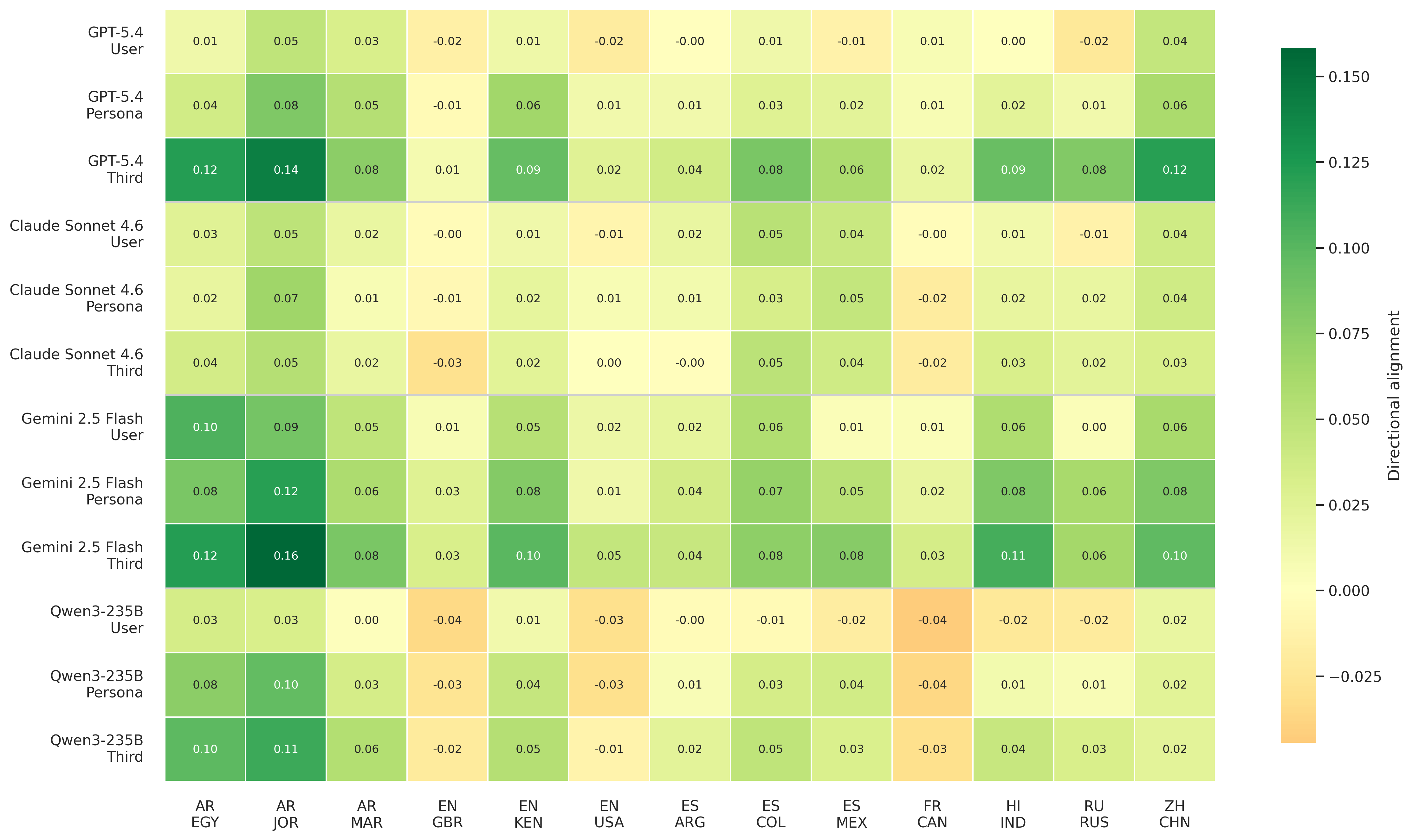}
\caption{Directional alignment by language-country slice and prompt condition for the four hosted models. We observe that the countries for which directional alignment is strongest—including Arabic-speaking countries, India, and China—are relatively consistent across models.}
\label{fig:country-alignment}
\end{figure*}

\begin{figure*}[!t]
\centering
\includegraphics[width=\textwidth]{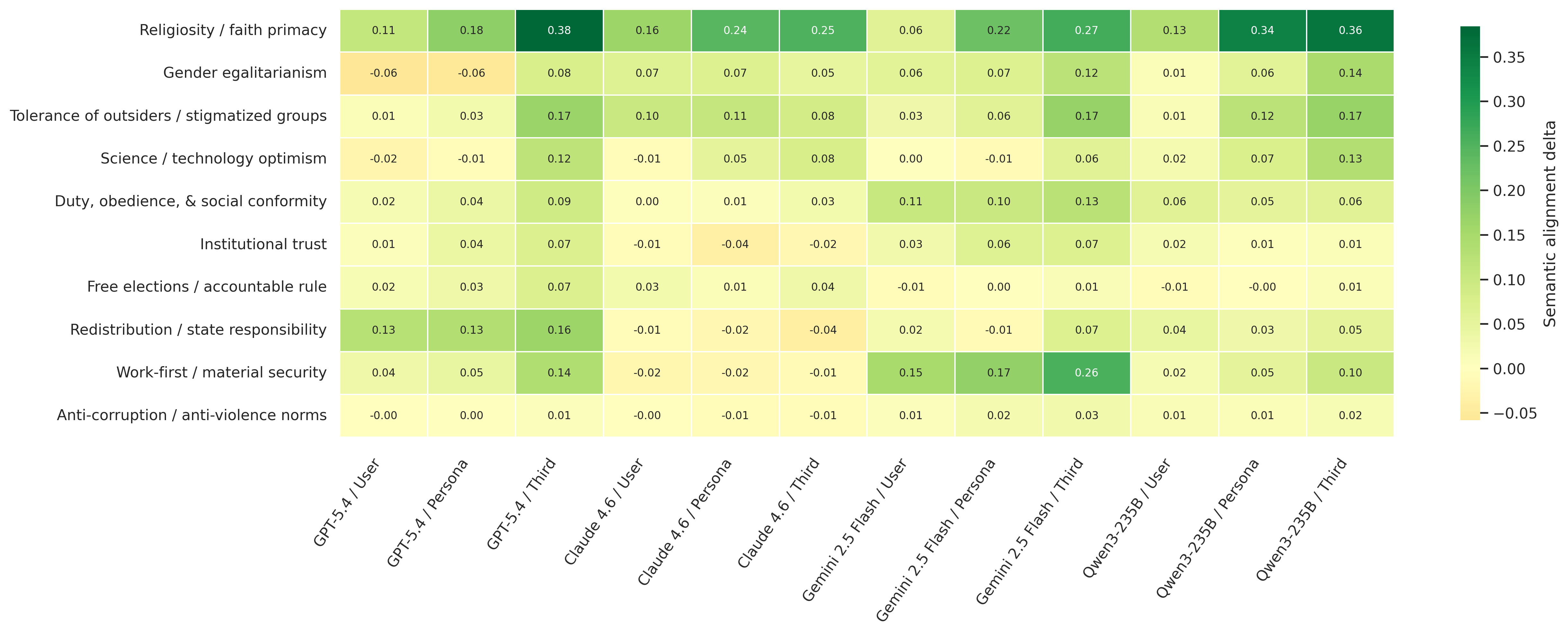}
\caption{Semantic-axis alignment by prompt variant and hosted model. Certain semantic axes, especially religiosity / faith primacy, are consistently strong across all models. Others vary more substantially by model: GPT-5.4 is a high-end outlier for redistribution / state responsibility, for example, as is Gemini 2.5 Flash for work-first / material security.}
\label{fig:semantic-axis-heatmap}
\end{figure*}

\begin{figure*}[!t]
\centering
\includegraphics[width=0.9\textwidth]{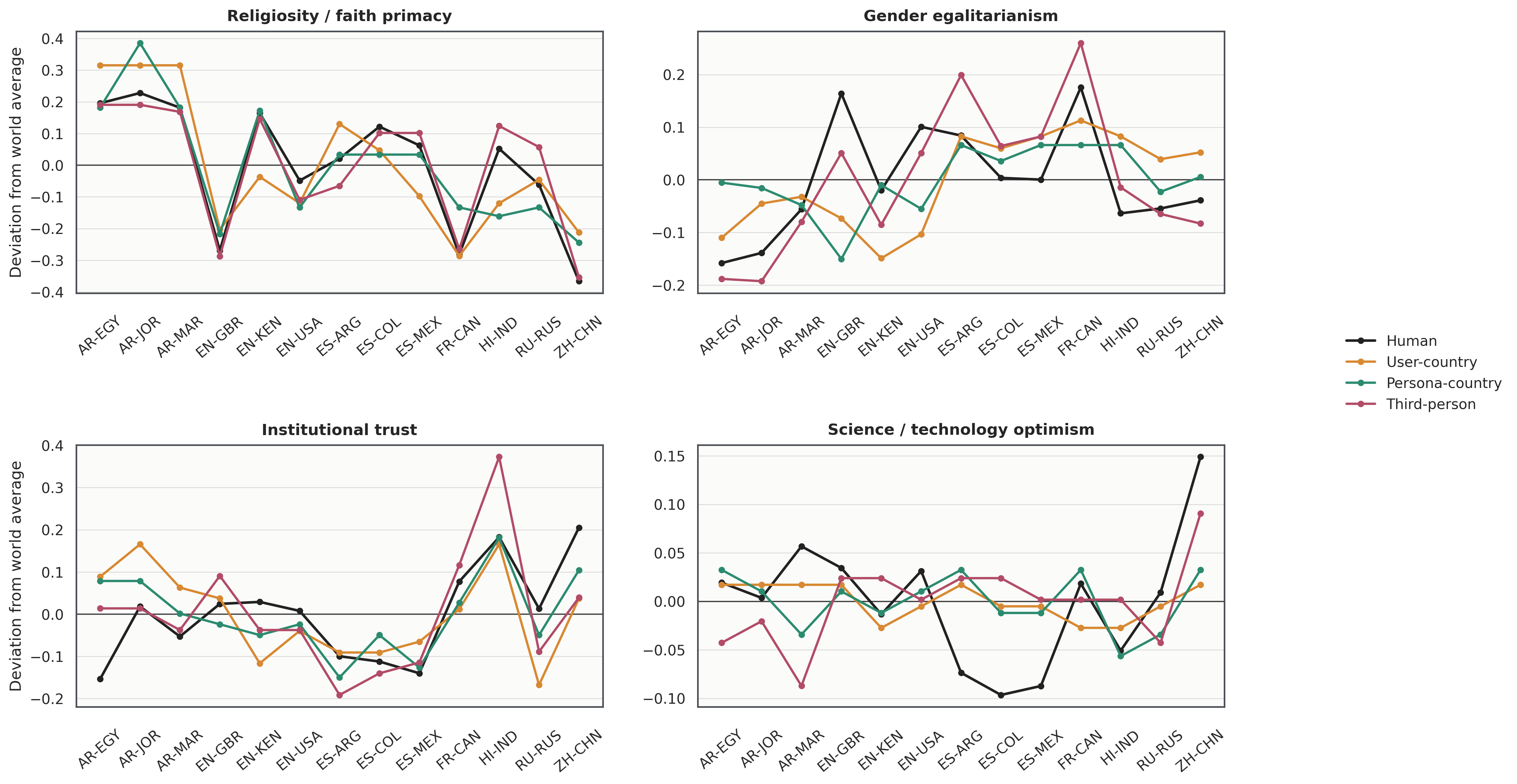}
\caption{Country deviations on representative semantic axes, centered on the world-average human reference. All LLM data in this figure is for GPT-5.4. This figure gives a more detailed country-level view of semantic-axis alignment, including notable anomalies such as GPT-5.4's overestimation of science / technology optimism for Latin American countries.}
\label{fig:country-axis-panels}
\end{figure*}

\section{Results}

\subsection{Overall Prompt Comparison}

Figure~\ref{fig:overall-metrics} reports the main benchmark summary for the four hosted models. Coverage represents the proportion of questions with recoverable answers, excluding cases in which the model hedged or refused to answer. \textbf{The main overall result is that country conditioning usually improves directional alignment relative to the language-only prompt, and third-person forecasting is the strongest prompt condition for all hosted models except Claude Sonnet 4.6.} Qwen's user-country variant is the main exception among country-conditioned prompts, with slightly negative directional alignment.

The strongest third-person alignment number is Gemini's $0.075$, narrowly above GPT-5.4's $0.073$. However, Gemini reaches this score with much lower coverage: $79.0\%$ under third-person prompting versus GPT-5.4's $96.7\%$, due to its high refusal rates on certain languages (see Appendix~\ref{app:figures-tables}, Figure~\ref{fig:refusal-patterns}). Qwen differs from GPT and Gemini: its vanilla and user-country prompts yield negative alignment, but third-person prompting lifts it to a clearly positive $0.034$. Claude shows the weakest prompt separation, with its country-conditioned variants clustered tightly between $0.018$ and $0.019$.

\subsection{Forecasting Versus Personas}

The prompt comparison suggests that third-person prompting changes the task being solved. User-country and persona-country prompts frame the answer as something like ``LLM response under a country cue,'' whereas the third-person prompt asks for an explicit forecast about a respondent from the target country. This distinction appears at the slice level: \textbf{third-person directional alignment exceeds persona-country alignment in all 13 language-country slices for GPT-5.4 and Gemini 2.5 Flash, in 11 of 13 slices for Qwen, and in 7 of 13 slices for Claude.}

The full slice matrix in Figure~\ref{fig:country-alignment} shows that this advantage varies geographically. The same countries recur near the top across models: Jordan is the best third-person slice for GPT-5.4, Gemini, and Qwen, and Egypt, India, China, Kenya, and Colombia also rank consistently high. These cases have country-conditioned targets that differ visibly from the baseline on high-salience social values, religiosity, or security-related tradeoffs.

By contrast, Great Britain, Canada, and the United States are weak slices for every model, especially Qwen and Claude. For GPT-5.4, third-person prompting still improves alignment in these English/French North Atlantic slices, but only slightly: the gains are $0.010$ for Great Britain, $0.018$ for Canada, and $0.025$ for the United States. Qwen is more revealing: its third-person scores are negative for all three ($-0.020$, $-0.030$, and $-0.012$, respectively), despite nontrivial shift magnitudes. On these country slices, only about a quarter of scorable items improve while about 40\% worsen, with repeated misses concentrated in trust, democracy/governance, corruption, and related institutional questions.

\subsection{Country-Specific Effects}

The country slices with strongest directional alignment tend to have human WVS profiles that depart from Anglophone defaults along several mutually reinforcing dimensions at once. In aggregate, Arabic slices combine higher religiosity and more traditional social attitudes; India combines stronger religiosity with more work-first and authority-oriented responses; China combines distinctive institutional and social patterns; and Colombia and Kenya show marked differences on several value-laden items. \textbf{In these cases, the models appear able to retrieve a coarse but directionally useful cultural schema.}

Conversely, Great Britain, Canada, and the United States appear to be hard cases precisely because the needed adjustment is smaller and more selective. Forecasting prompts still induce substantial movement in those slices, but the movement often buys little alignment, suggesting that the models handle large, legible cross-country contrasts more easily than fine-grained distinctions among relatively similar countries.

Figure~\ref{fig:country-alignment} also shows that this pattern varies partly by model. Gemini retains positive alignment even in its weakest slices, while Qwen's failures concentrate much more sharply in the Anglophone and Canadian slices. Still, the overall ordering of strong and weak country slices remains notably consistent across models.

\subsection{Where The Effects Concentrate}

Prompt effects vary across categories. Under third-person prompting, GPT-5.4 and Gemini both have positive directional alignment in all seven broad categories. GPT-5.4 is strongest in Religion and Science ($0.187$) and Economic Fairness, Work, and Material Conditions ($0.111$), and Morality and Worldview ($0.076$). Gemini shows a similar but sharper concentration in Religion and Science ($0.175$), Economic Fairness, Work, and Material Conditions ($0.130$), and Public Order, Corruption, Migration, and Security ($0.125$).

Qwen's strongest third-person categories are Religion and Science ($0.187$) and Gender, Equality, and Social Inclusion ($0.090$), but it is slightly negative on Democracy, Governance, and Public Authority ($-0.014$) and Trust, Civic Belonging, and Institutions ($-0.022$). Claude is also uneven, but joins the other models in having Religion and Science as its clearest positive category.

At the question level, the strongest alignment gains concentrate in items such as importance of God, belief in God, freedom versus security, environment versus growth, and worries about war or terrorism. Weak or negative items cluster more around institutional trust, election fairness, corruption accountability, and some democracy-essential questions. \textbf{The broad pattern is that models track country differences more readily on socially legible value questions than on questions requiring finer-grained institutional modeling.}

\subsection{Semantic Direction Is Axis-Specific}

The semantic-axis results argue against a one-dimensional account such as ``the models have more traditional values under country prompting.'' Some axis-level changes fit that description, especially on religiosity and some gender-role questions, but the effect is not uniform across the ten-axis inventory.

Across all four models, religiosity / faith primacy is the easiest axis. GPT-5.4 has the largest third-person delta there ($0.384$), while Qwen ($0.360$) and Gemini ($0.266$) also move strongly in the same direction. Figure~\ref{fig:semantic-axis-heatmap} shows that the second-tier strengths differ by model. GPT-5.4 is the clear outlier on economic interventionism ($0.165$), far ahead of Gemini ($0.067$), Qwen ($0.053$), and Claude ($-0.037$). This suggests that GPT is unusually good at reorienting on redistribution and state-responsibility questions, rather than only on socio-cultural questions.

Gemini's most distinctive strength is productivist materialism / work-first orientation ($0.256$), where it substantially outperforms all other models. Qwen, meanwhile, improves more than Claude on gender egalitarianism and outgroup inclusion, but remains much weaker on institutional trust and democratic pluralism. \textbf{The semantic-axis layer further supports an axis-specific interpretation: prompting helps most where cross-country differences are socially legible and less where they depend on more specific institutional knowledge.}

The country-by-axis panels in Figure~\ref{fig:country-axis-panels} present a case study for GPT-5.4, comparing deviations from the world-average human reference across four axes. On religiosity / faith primacy, GPT broadly recovers the human ordering of countries: the Arabic slices sit well above the world average, while China, Canada, and Great Britain sit well below it. On gender egalitarianism, the model is directionally reasonable but not perfectly calibrated: it overstates egalitarian deviations for Argentina and Canada and understates them for Great Britain.

Institutional trust is the clearest failure mode. Under third-person prompting, GPT assigns India much more above-average trust than the human reference does, misses China's strongly above-average trust signal, and softens several low-trust countries toward the middle. Science / technology optimism is mixed in a different way: GPT is fairly close for most of the panel, but it systematically overestimates Latin American optimism, predicting Argentina and Colombia as slightly above the world average even though the human reference is below it, while also pushing Morocco below average despite a modestly positive human deviation.

\subsection{Localized Models}
\label{sec:localized}

We also compare the hosted models to three narrower localized models on corresponding language slices: Jais-2-8B-Chat for Arabic, YandexGPT-5-Lite-8B for Russian, and Airavata for Hindi \cite{sengupta2023jaisjaischatarabiccentricfoundation, gala2024airavataintroducinghindiinstructiontuned}. \textbf{None of these localized models exceeds GPT-5.4 or Gemini on the overlapping third-person slice summaries, although Airavata is competitive with Claude and Qwen on Hindi.}

The slice-matched comparison is nonetheless informative. Airavata is the strongest localized baseline: on Hindi/India, its third-person directional alignment is $0.031$, essentially tied with Claude's $0.032$, though still below Qwen ($0.042$), GPT-5.4 ($0.093$), and Gemini ($0.108$). YandexGPT-5-Lite-8B is respectable on Russian in persona-country mode ($0.020$), where it slightly exceeds GPT-5.4 and Qwen, but its third-person score falls to $0.010$, leaving it below all four hosted models. Jais is the weakest case: its Arabic third-person average alignment is negative ($-0.022$), even though the hosted models are all positive on the same slices. Appendix~\ref{app:localized} gives the slice-by-slice comparison in more detail.

\section{Conclusion}
This paper shows that personalization, personas, and forecasting are not interchangeable interfaces for eliciting value judgments from LLMs. Across WVS questions, country-conditioned prompts often change model responses substantially, but the direction and usefulness of that change depend on the framing: third-person forecasting most consistently moves responses toward matched human distributions, while user-country and persona-country prompts produce weaker or less predictable alignment. The strongest successes appear on socially legible axes such as religiosity and gender roles, while institutional trust and democratic values expose persistent gaps. These findings make prompt framing a core methodological choice for cultural alignment work. Future evaluations should report and justify the modality they use, rather than treating identity-conditioned prompts as equivalent.

\bibliography{santa_icml}
\bibliographystyle{icml2026}

\newpage
\appendix

\section{Question Selection Details}
\label{app:selection}
The final benchmark subset contains 101 standalone WVS question units. Selection followed four main rules:
\begin{enumerate}
\item prefer wording that is portable across countries and does not depend on country-specific party lists, institution rosters, or local annexes;
\item prefer questions that remain interpretable as standalone prompts;
\item exclude self-evaluative or immediate-local-context items such as life satisfaction, happiness, self-rated health, and trust in family;
\item keep substantively useful controversial items, but remove near-duplicates.
\end{enumerate}

The three grouped selection units are:
\begin{itemize}
\item \texttt{SU\_CHILD\_QUALITIES} for \texttt{Q7-Q17};
\item \texttt{SU\_NEIGHBORS} for \texttt{Q18-Q26};
\item \texttt{SU\_IMMIGRATION\_EFFECTS} for \texttt{Q122-Q129}.
\end{itemize}
These grouped units are rendered with one response line per member item and are averaged back to a single parent score during evaluation. Selection operated on standalone units rather than raw codebook rows because several WVS batteries are experienced by respondents as one question with multiple recorded members.

\section{Prompt Templates}
\label{app:prompts}
Each prompt consists of four pieces: an optional country preamble, the localized question text, the localized response options, and a language-specific answer suffix. The vanilla prompt uses no country preamble.

\begin{table*}[h]
\caption{Country-prompt preambles by language. \texttt{\{country\_name\}} is replaced with the localized country name for the target slice.}
\label{tab:prompt-preambles}
\centering
\small
\begin{tabular}{p{0.08\textwidth}p{0.28\textwidth}p{0.28\textwidth}p{0.28\textwidth}}
\toprule
Lang. & User-country & Persona-country & Third-person \\
\midrule
Arabic & \foreignlanguage{arabic}{أنا من} \{country\_name\}. & \foreignlanguage{arabic}{أجب كما لو أنك من} \{country\_name\}. & \foreignlanguage{arabic}{كيف تعتقد أن شخصًا من} \{country\_name\} \foreignlanguage{arabic}{سيجيب؟} \\
English & I am from \{country\_name\}. & Answer as if you are from \{country\_name\}. & How do you think someone from \{country\_name\} would answer? \\
Spanish & Soy de \{country\_name\}. & Responde como si fueras de \{country\_name\}. & ¿Cómo crees que respondería alguien de \{country\_name\}? \\
French & Je viens de \{country\_name\}. & Répondez comme si vous veniez de \{country\_name\}. & Comment pensez-vous qu'une personne de \{country\_name\} répondrait ? \\
Hindi & \foreignlanguage{hindi}{मैं} \{country\_name\} \foreignlanguage{hindi}{से हूँ।} & \foreignlanguage{hindi}{उत्तर ऐसे दें जैसे आप} \{country\_name\} \foreignlanguage{hindi}{से हों।} & \foreignlanguage{hindi}{आपके विचार से} \{country\_name\} \foreignlanguage{hindi}{का कोई व्यक्ति कैसे उत्तर देगा?} \\
Russian & \foreignlanguage{russian}{Я из} \{country\_name\}. & \foreignlanguage{russian}{Отвечайте так, как если бы вы были из} \{country\_name\}. & \foreignlanguage{russian}{Как, по-вашему, ответил бы человек из} \{country\_name\}? \\
Chinese & 我来自\{country\_name\}。 & 请像你来自\{country\_name\}一样回答。 & 你认为来自\{country\_name\}的人会怎样回答？ \\
\bottomrule
\end{tabular}
\end{table*}

\begin{table*}[h]
\caption{Answer suffixes by language. This suffix follows the localized question text and response options in every prompt condition.}
\label{tab:prompt-suffixes}
\centering
\small
\begin{tabular}{p{0.08\textwidth}p{0.84\textwidth}}
\toprule
Lang. & Answer suffix \\
\midrule
Arabic & \foreignlanguage{arabic}{أجب باستخدام أحد خيارات الإجابة المعروضة فقط. إذا لم تكن متأكدًا تمامًا، فاختر أقرب إجابة تعكس رأيك.} \\
English & Answer using only one of the listed response options. If you are not fully sure, choose the closest option that best matches your view. \\
Spanish & Responde usando solo una de las opciones de respuesta indicadas. Si no estás del todo seguro, elige la opción más cercana a tu opinión. \\
French & Répondez en utilisant uniquement l'une des options proposées. Si vous n'êtes pas entièrement sûr, choisissez l'option la plus proche de votre opinion. \\
Hindi & \foreignlanguage{hindi}{दिए गए उत्तर विकल्पों में से केवल एक का उपयोग करके उत्तर दें। यदि आप पूरी तरह निश्चित नहीं हैं, तो अपनी राय के सबसे करीब विकल्प चुनें।} \\
Russian & \foreignlanguage{russian}{Отвечайте, используя только один из предложенных вариантов ответа. Если вы не вполне уверены, выберите вариант, который ближе всего к вашей позиции.} \\
Chinese & 请只使用给出的一个回答选项作答。如果你不能完全确定，就选择最接近你看法的选项。 \\
\bottomrule
\end{tabular}
\end{table*}

\section{Additional Metric and Aggregation Details}
\label{app:metrics}
Human response distributions are computed from WVS respondent weights over substantive responses only. For a given language or language-country group, the distribution mass on each response option is:
\begin{align*}
    P(\text{option } k) = \\ \frac{\text{weighted substantive responses choosing }k}{\text{total substantive response weight for the group}}.
\end{align*}

These option shares sum to 1 within each question member. Missing, blank, and other non-substantive responses are excluded from this distribution and reported through coverage instead.

Bundled questions are scored in three steps:
\begin{enumerate}
\item compute the metric separately for each member question;
\item average member scores back to the parent selection unit;
\item average parent selection units with equal weight in aggregate summaries.
\end{enumerate}
This prevents larger batteries from dominating the benchmark solely because they contain more recorded members.

\section{Semantic Axis Inventory}
\label{app:axes}
\Cref{tab:axis-inventory} lists the ten semantic axes used for direction-of-shift analysis. Each mapped question-member receives a semantic value in $[0,1]$, oriented so that larger values always mean more of the positive pole. The final column gives the mapped benchmark units for each axis, using WVS question IDs and benchmark selection-unit IDs for the bundled batteries.

\begin{table*}[h]
\caption{Semantic-axis inventory.}
\label{tab:axis-inventory}
\centering
\footnotesize
\begin{tabular}{p{0.25\textwidth}p{0.20\textwidth}p{0.20\textwidth}p{0.25\textwidth}}
\toprule
\textbf{Axis} & \textbf{Positive pole} & \textbf{Negative pole} & \textbf{Mapped question units} \\
\midrule
Gender egalitarianism & gender equality & gender hierarchy & \texttt{Q28, Q29, Q30, Q31, Q32, Q33, Q35, Q36, Q182, Q189, Q233, Q249} \\
\midrule
Outgroup inclusion & tolerance of outsiders and stigmatized groups & exclusion and social distance & \texttt{Q36, Q121, Q130, Q182, SU\_NEIGHBORS} \\
\midrule
Religiosity / faith primacy & religiosity and deference to faith & secular orientation & \texttt{Q160, Q164, Q165, Q169, Q173, SU\_CHILD\_QUALITIES} \\
\midrule
Science / technology optimism & science and technology seen as beneficial & science and technology skepticism & \texttt{Q44, Q158, Q161, Q162, Q163} \\
\midrule
Traditional authority & duty, obedience, deference, conformity & autonomy, self-expression, reformism & \texttt{Q37, Q38, Q42, Q45, Q152, Q154, Q156, Q235, Q237, SU\_CHILD\_QUALITIES} \\
\midrule
Civic integrity & anti-corruption and anti-violence norms & tolerance of corruption, theft, or violence & \texttt{Q112, Q116, Q120, Q178, Q179, Q180, Q181, Q189, Q191, Q192, Q194} \\
\midrule
Economic interventionism & redistribution and state responsibility & market individualism and inequality tolerance & \texttt{Q106, Q107, Q108, Q149, Q241, Q244} \\
\midrule
Productivist materialism & work-first, growth-first, order, material security & post-material and expressive priorities & \texttt{Q39, Q40, Q41, Q109, Q110, Q111, Q150, Q152, Q154, Q156} \\
\midrule
Institutional trust & confidence in institutions and social trust & institutional distrust and suspicion & \texttt{Q57, Q64, Q65, Q66, Q68, Q69, Q70, Q71, Q72, Q74, Q75, Q76, Q78} \\
\midrule
Democratic pluralism & free elections, rights, accountable democratic rule & authoritarian or securitarian acceptance & \texttt{Q149, Q150, Q196, Q197, Q198, Q224, Q225, Q227, Q228, Q229, Q230, Q231, Q232, Q235, Q237, Q238, Q241, Q243, Q244, Q246, Q250, Q253} \\
\bottomrule
\end{tabular}
\end{table*}

\section{Parser and Coverage Notes}
\label{app:parser}
We use deterministic answer extraction that prioritizes explicit answer-bearing text, including numbered responses and exact option labels, before applying conservative fallback matching. Hedge and refusal flags are tracked separately from parsing status. For bundled rows, the primary benchmark score requires all expected members to be parsed.

\section{Localized Models}
\label{app:localized}

We compare three slice-matched localized baselines to the four hosted models on the subsets for which a region-specific model is available: Jais-2-8B-Chat on the three Arabic slices (Egypt, Jordan, Morocco), YandexGPT-5-Lite-8B on Russian/Russia, and Airavata on Hindi/India. The comparison uses the same prompt conditions and the same directional-alignment metric as in the main benchmark, but aggregates only over the overlapping slice or slice family.

The localized models do not dominate the hosted systems on their home slices. For Arabic, Jais is weaker than all four hosted models under every conditioned prompt. Its user-country average over the three Arabic slices is modestly positive ($0.011$), its persona-country average is near zero ($0.004$), and its third-person average is negative ($-0.022$). By contrast, the hosted models are all positive on Arabic third-person prompting, ranging from $0.036$ for Claude to $0.121$ for Gemini, with GPT-5.4 at $0.114$ and Qwen at $0.088$. Jais also has substantially lower coverage than the strongest hosted systems, with overall conditioned coverage between $79.5\%$ and $84.5\%$.

Airavata is the most competitive localized baseline. On Hindi/India it reaches $0.023$ for user-country, $0.031$ for persona-country, and $0.031$ for third-person prompting, with roughly $89\%$--$91\%$ coverage across prompt conditions. These numbers place it close to Claude on the same slice and above GPT-5.4 in user-country mode, but still clearly below GPT-5.4 and Gemini under third-person prompting. Airavata therefore looks like a credible localized baseline rather than a trivial failure case, but it does not overturn the main finding that the strongest hosted models are better respondent forecasters.

YandexGPT-5-Lite-8B performs respectably on Russian. Its persona-country score on Russia is $0.020$, slightly above GPT-5.4 ($0.012$) and Qwen ($0.006$), and close to Claude ($0.018$), while maintaining essentially complete coverage. But its third-person score is only $0.010$, which leaves it well below GPT-5.4 ($0.081$) and Gemini ($0.064$), and below Claude ($0.023$) and Qwen ($0.032$) as well. The localized models are therefore best understood as selective competitors on some prompt styles and slices, not as across-the-board replacements for the strongest general-purpose systems. Figure~\ref{fig:sovereign} visualizes these slice-matched comparisons.

\section{Additional Figures and Tables}
\label{app:figures-tables}

\begin{figure*}[h]
\centering
\includegraphics[width=\textwidth]{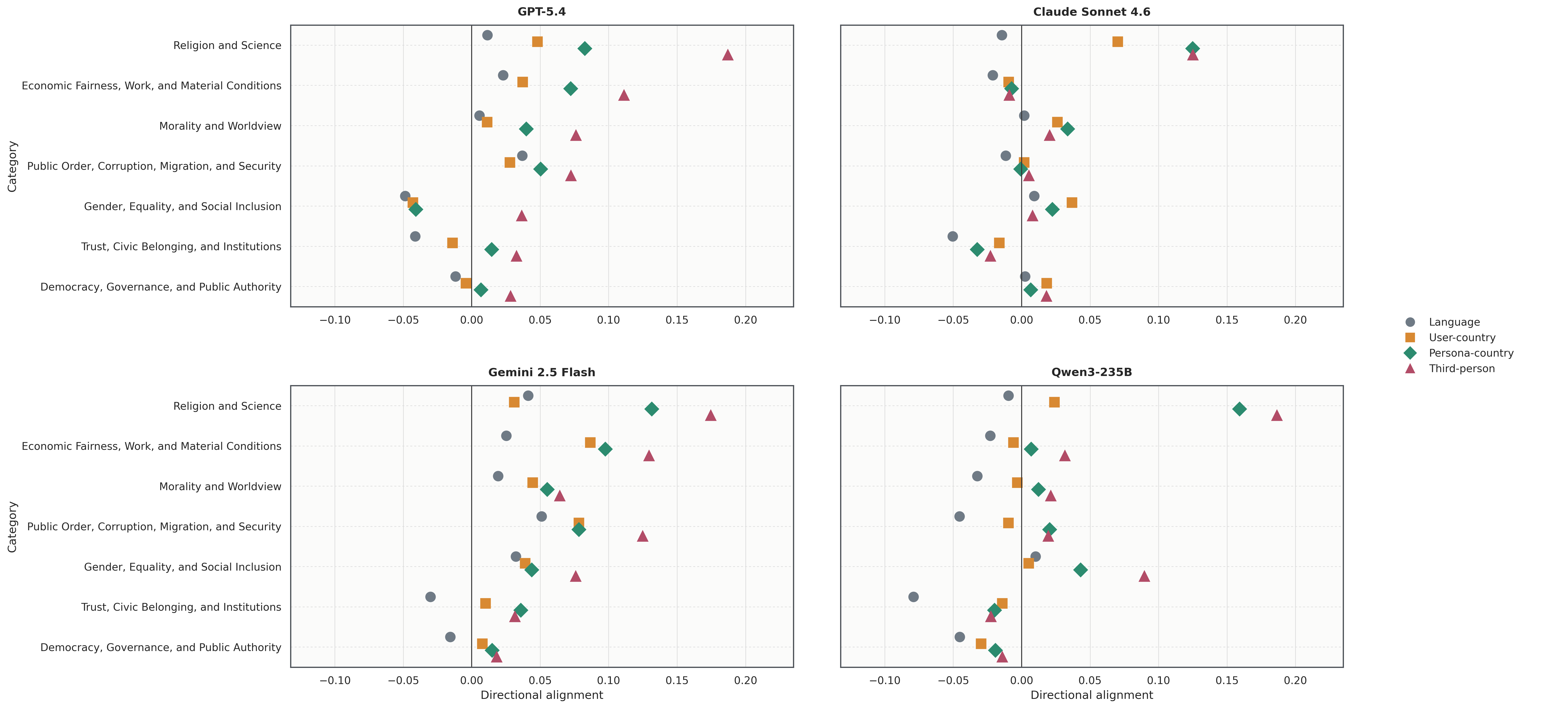}
\caption{Category-level directional alignment for both models and prompt variants.}
\label{fig:category-alignment}
\end{figure*}

These figures serve as diagnostics and decompositions for the main benchmark results. Figure~\ref{fig:category-alignment} shows that aggregate gains are carried disproportionately by a subset of categories, especially Religion and Science and, for GPT-5.4 and Gemini, Economic Fairness, Work, and Material Conditions.

\begin{figure*}[h]
\centering
\includegraphics[width=\textwidth]{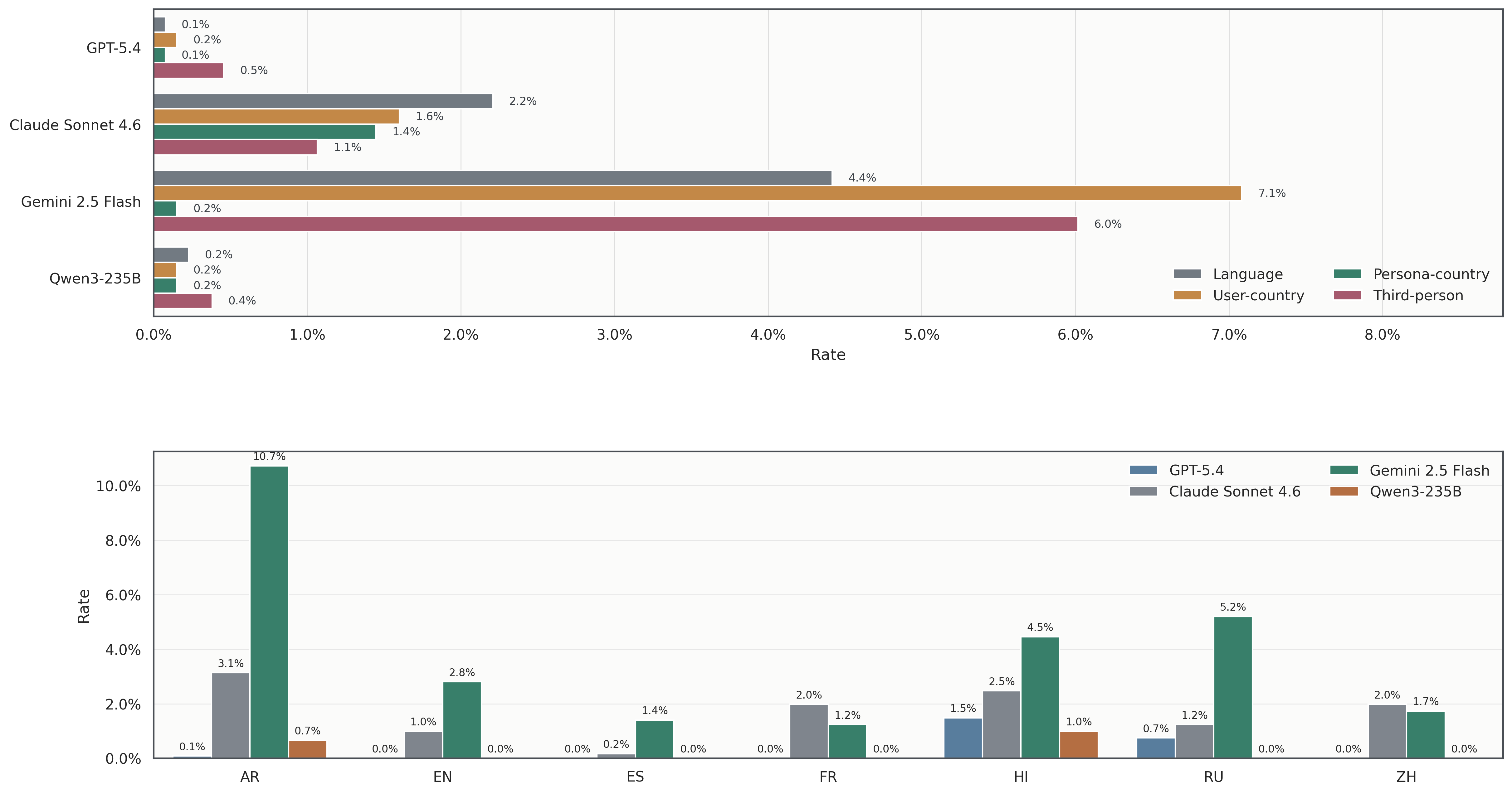}
\caption{Prompt-level and language-level refusal patterns for the four hosted models.}
\label{fig:refusal-patterns}
\end{figure*}

Figure~\ref{fig:refusal-patterns} is a coverage diagnostic. It shows that the models differ not only in alignment but also in whether they will answer the question in the first place. GPT-5.4 and Qwen have near-zero refusal rates across most settings, Claude has low but nonzero refusal rates concentrated in Arabic, and Gemini's weaker effective coverage is driven largely by conditioned-prompt refusals, especially in Arabic and to a lesser extent Russian and Hindi.

\begin{figure*}[h]
\centering
\includegraphics[width=\textwidth]{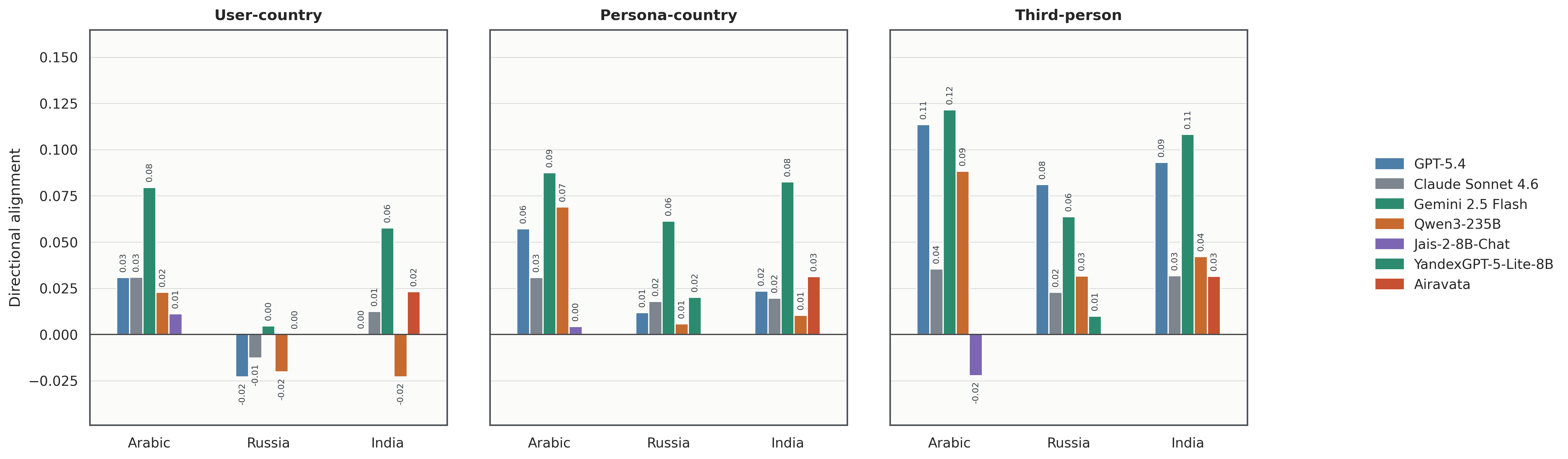}
\caption{Hosted versus localized models on the Arabic, Russian, and Hindi slice-matched comparisons.}
\label{fig:sovereign}
\end{figure*}

Figure~\ref{fig:sovereign} gives the localized-model comparison used in Section~\ref{sec:localized}. The important reading is slice-relative rather than global: each localized model is compared only on the language-country cells for which it is available. The figure makes clear that Airavata is the closest localized competitor, Yandex is competitive mainly in persona-country prompting, and Jais does not recover the Arabic forecasting gains seen in the hosted systems.

\section{Ethics, Broader Impact, and Reproducibility}

This benchmark is descriptive rather than normative. A positive alignment score means that a model moved toward the observed WVS response distribution under our prompting and scoring setup; it does not mean that the model is ``correct'' in any broader moral or political sense, nor that the matched survey responses should be treated as a social ideal. The benchmark is also about aggregate distributions, not about any individual's beliefs. Even when a country-level forecast is accurate on average, it should not be used to stereotype or essentialize respondents from that country.

The work also poses a familiar risk in cultural modeling: prompting a model to answer ``as if'' it were from a given country can encourage the use of coarse cultural heuristics. Some of those heuristics may help with forecasting, but they may also encode flattening stereotypes, especially on socially sensitive dimensions such as religion, gender roles, or attitudes toward minorities. For that reason, we report axis-specific successes and failures rather than presenting country prompting as a general solution. In particular, our results show that models can look strong on salient value dimensions while remaining weak on institutional trust, democratic process, and other subtler constructs.

The underlying human data come from the World Values Survey, and our benchmark uses weighted aggregate response distributions rather than individual-level prediction targets. Reproducibility is supported by deterministic scoring rules, explicit prompt templates, fixed question selection, and explicit aggregation procedures, all described in this paper. The appendix tables and figures are computed from the same benchmark runs discussed in the main text, and the scoring and aggregation rules used to produce them are specified in Section~\ref{app:metrics}. The main remaining source of irreproducibility is the model side itself: hosted LLM outputs may change over time, so exact benchmark values should be interpreted as tied to the specific model snapshots and runs analyzed here.

\end{document}